\documentclass[conference]{IEEEtran}
\IEEEoverridecommandlockouts

\usepackage{cite}
\usepackage{amsmath,amssymb,amsfonts}
\usepackage{algorithmic}
\usepackage{graphicx}
\usepackage{textcomp}
\usepackage{xcolor}
\usepackage{multirow}
\usepackage{float}
\usepackage{xurl} 
\usepackage{hyperref}
\hypersetup{breaklinks=true}
\def\BibTeX{{\rm B\kern-.05em{\sc i\kern-.025em b}\kern-.08em
    T\kern-.1667em\lower.7ex\hbox{E}\kern-.125emX}}
\begin{document}

\title{A Domain-Agnostic Neurosymbolic Approach for Big Social Data Analysis: Evaluating Mental Health Sentiment on Social Media during COVID-19\\
\thanks{Research is funded as part of NSF Awards: "Spokes: MEDIUM: MIDWEST: Collaborative: Community-Driven Data Engineering for Substance Abuse Prevention in the Rural Midwest" (Award\#: 1956009), "EAGER: Advancing Neurosymbolic AI with Deep Knowledge-infused Learning" (Award\#: 2133842), and "EAGER: Knowledge-guided neurosymbolic AI with guardrails for safe virtual health assistants" (Award\#: 2335967)}
}

\author{
    Vedant Khandelwal\(^{1}\), Manas Gaur\(^{2}\), Ugur Kursuncu\(^{3}\), Valerie L. Shalin\(^{1,4}\), Amit P. Sheth\(^{1}\) \\
    \(^{1}\)AI Institute, University of South Carolina, Columbia, South Carolina \\
    \(^{2}\)Department of Computer Science and Engineering, University of Maryland, Baltimore County, Maryland \\
    \(^{3}\)Institute for Insight, Georgia State University, Atlanta, Georgia \\
    \(^{4}\)Department of Psychology, Wright State University, Dayton, Ohio \\
    vedant@email.sc.edu, manas@umbc.edu, ugur@gsu.edu, valerie.shalin@wright.edu, amit@sc.edu
}

\maketitle

\begin{abstract}
Monitoring public sentiment via social media is potentially helpful during health crises such as the COVID-19 pandemic. However, traditional frequency-based, data-driven neural network-based approaches can miss newly relevant content due to the evolving nature of language in a dynamically evolving environment. Human-curated symbolic knowledge sources, such as lexicons for standard language and slang terms, can potentially elevate social media signals in evolving language. We introduce a neurosymbolic method that integrates neural networks with symbolic knowledge sources, enhancing the detection and interpretation of mental health-related tweets relevant to COVID-19. Our method was evaluated using a corpus of large datasets (approximately 12 billion tweets, 2.5 million subreddit data, and 700k news articles) and multiple knowledge graphs. This method dynamically adapts to evolving language, outperforming purely data-driven models with an F1 score exceeding 92\%. This approach also showed faster adaptation to new data and lower computational demands than fine-tuning pre-trained large language models (LLMs). This study demonstrates the benefit of neurosymbolic methods in interpreting text in a dynamic environment for tasks such as health surveillance.
\end{abstract}


\section{Introduction}
Online platforms like X (formerly Twitter) are vital for capturing real-time public sentiment, especially during crises. With X generating approximately 500 million tweets daily \cite{tweetstat}, researchers have a substantial source of data to monitor public discourse \cite{Sheth_conversation, weger2023trends, wang2023global, thakur2022large}. This capability is particularly crucial in contexts like the COVID-19 pandemic, which has significantly impacted mental health, evidenced by marked increases in anxiety, depression, and substance use \cite{WHO_2022}. However, most existing studies on the mental health analysis of social media data have been retrospective, limiting their effectiveness in providing timely insights. There is a pressing need for near-real-time analysis tools to swiftly detect and respond to emerging trends in big data, thereby enhancing the efficacy of public health interventions and policy responses.

Despite the utility of neural network-based techniques such as large language models (LLMs), attention models, and word embedding models like Word2Vec \cite{mikolov2013efficient} and GloVe \cite{pennington2014glove}, their application to dynamic social media language poses challenges. These models struggle with the flexibility needed to adapt to rapid changes in language, such as the emergence of terms like ``Zoom fatigue" \cite{jovanovic2024trends}, which gained relevance during the pandemic as remote work became common. The reliance on vast computational resources further complicates rapid adaptation to new linguistic phenomena, as LLMs and other traditional models depend heavily on historical data. This limitation makes them less effective for real-time monitoring of evolving terms and trends crucial in health monitoring contexts. A neurosymbolic approach offers a more adaptable, efficient framework that integrates evolving linguistic data.

To manage the rapid evolution of online language effectively, our neurosymbolic approach uniquely combines a domain-specific language model with extensive integration of several domain-specific and general-purpose knowledge bases (KBs). This integration facilitates the dynamic incorporation of emerging vocabulary and contextual shifts, enabling near-real-time analysis of online discourse. Utilizing a vast dataset of approximately 12 billion tweets, 2.5 million subreddit posts, and 700k news articles, our method leverages the Zero-Shot Semantic Encoding and Decoding Optimization (SEDO) framework—initially designed for image processing \cite{kodirov2017semantic} and adapted here for mental health applications in social media text\cite{gaur2018let}. SEDO calculates relevance scores by evaluating the semantic similarity between new terms and established knowledge concepts, enhancing content representation, and enabling dynamic adaptation to context-dependent meanings. This method significantly improves over the less precise soft match approach previously used \cite{gaur2018let}. Additionally, our approach integrates multiple knowledge sources for managing semantic gaps \cite{bajaj2022knowledge}, and its scalable design makes it applicable to a broad range of domains beyond mental health.

Our neurosymbolic approach outperforms traditional data-driven models with an F1 score of $>92\%$, demonstrating adaptability and efficiency. It requires fewer computational resources than large language models. Additional experiments demonstrate that our approach excels compared to pre-trained LLMs. Further validation through an ablation study reveals the contribution of components to the overall performance of the method. Additionally, a triangulation study affirms the model's adaptability, showing practical application to new, previously unseen datasets, thereby confirming its broader applicability.

This paper is structured according to the progressive development and validation of our neurosymbolic approach. Following this introduction, a background and literature review presents the relevant theories and existing works. Our methods section elaborates on integrating symbolic AI with neural networks and details the adaptive techniques employed. The experimental setup and results sections then present the data sources, preparation processes, and the outcomes of our studies, including ablation and triangulation analyses. Finally, the discussion encapsulates the broader implications of our findings, assesses the advantages of our approach, and suggests avenues for future research.

\section{Background and Literature Review}

Several analytical methods apply to studying social media data: traditional word-based analysis, topic modeling, embeddings, and language models. We describe these, particularly noting the shortcomings in capturing the complex nature of mental health discourse.
We contrast these limitations with the advantages of employing knowledge bases, deep learning, and knowledge-infused learning. Our literature review also includes studies on adaptive learning processes and neurosymbolic methods suitable for (near) real-time analytics with large social data volumes. 

\subsection{Word-based Analysis and Topic Modelling}

Traditional domain-specific analyses of public discourse often utilize word-based methods, such as keyword tracking and topic modeling.
These methods have provided valuable insights into public sentiment and social support dynamics \cite{arora2021role, colella2023mental, esener2023seeking}, as well as identifying trends such as increased anxiety \cite{tankut2022analysis,osakwe2021identifying}, depression \cite{sadasivuni2022timestamp, santarossa2022understanding}, and loneliness \cite{koh2022loneliness}. However, they exhibit significant limitations, particularly in adapting to dynamically evolving languages like hashtags or new terms such as "Blursday" and "doomscrolling" \cite{russell2023gambling}. The reliance on predefined keywords can delay the identification of emerging trends and evolving language, potentially missing relevant mental health discourses \cite{thakur2022large}.

Topic modeling, particularly Latent Dirichlet Allocation (LDA), 
uncovers latent thematic structures in large text collections \cite{blei2003latent}. LDA models assess the co-occurrence of words within documents to determine topic distributions, which reveal the underlying thematic structures of the data. LDA can offer insights by treating phrases like ``social distancing'' and ``remote work'' as single n-gram units. Nevertheless, its reliance on a bag-of-words approach often overlooks critical linguistic structures such as word order and negation. For instance, the sentence ``I am not feeling well" could be misinterpreted by LDA as expressing wellness due to the presence of positive words like ``feeling" and ``well," ignoring the negation introduced by ``not."


\subsection{Knowledge Bases}

Knowledge bases (KBs) are expert-curated, structured compilations of interpretable symbolic concepts and relationships between them, representing the symbolic component of our neurosymbolic analysis. They embody ground truth conceptual knowledge and facilitate structured querying and systematic analysis \cite{krishna1992introduction}. Among various representations of KBs, ontologies are particularly instrumental in enhancing AI's ability to structure data to mirror human understanding and logical deduction. These frameworks situate concepts within an extensive framework of domain-specific knowledge that enhances interpretation \cite{bian2014knowledge, rawte2022tdlr}.

The Drug Abuse Ontology (DAO) \cite{lokala2022drug}, for example, provides descriptions of mental health conditions, extending beyond the lexical cues in a post. KBs resolve ambiguity in semantics and delineate relationships among concepts. While broad-spectrum KBs such as DBpedia and WikiData provide wide-ranging contexts that aid in interpreting entities like "isolation"—generally referred to as a medical protocol —domain-specific KBs such as UMLS (Unified Medical Language System) and SNOMED-CT (Systematized Nomenclature of Medicine--Clinical Terms) offer a mental health interpretation of "isolation" concerning restricted social interaction and loneliness without being physically ill.

KBs also facilitate inference and inheritance. For instance, using the DAO, if a tweet mentions symptoms like persistent sadness and loss of interest, a computational system can infer depression even if the term ``depression'' is not explicitly mentioned. Hierarchical relationships enable reasoning based on inheritance; for example, SNOMED-CT can help identify that phobias are a type of ``anxiety disorder,'' thus enabling the recognition of symptoms such as heart palpitations or insomnia. A semantic approach utilizing KBs has previously demonstrated acceptable accuracy \cite{perera2016implicit}.

However, a significant limitation of KBs is their slow update process, which can delay the integration of new information \cite{teniente1995updating}. To address this, our method constantly updates the lexicon with dynamic terms derived by utilizing these KBs and social media discourse, ensuring timely access to the latest data in rapidly changing environments. For instance, within the context of COVID-19, continuous updates to the lexicon from new terms obtained from subreddits, such as \textit{coronasomnia, quaranteam, quarantini, virtual happy hour, covid-19}, help maintain relevance.


\subsection{Embeddings and Neural Network Based Models}
Embeddings and neural network-based models 
convert words into high-dimensional vector representations that capture their semantic and syntactic attributes. Techniques like Word2Vec, GloVe, and fastText 
position semantically similar texts closely in vector space enhancing tasks such as sentiment analysis and machine translation \cite{mikolov2013efficient, pennington2014glove, joulin2017bag}. Cosine similarity can then gauge semantic proximity between vectors. While traditional embeddings provide a solid foundation by encapsulating local textual context, their representations sometimes fail to capture deeper contextual meanings necessary in specialized domains like mental health, such as distinguishing between expressions of mild sadness and clinical depression, where subtle linguistic meanings are critical for accurate detection and classification.

Deep learning models, including those based on complex neural networks and transformer architectures like GPT, extend these capabilities significantly. Incorporating multi-layered neural networks facilitates intricate pattern learning directly from data, enhancing performance across various NLP tasks such as sentiment analysis and named entity recognition. 
They 
provide improved understanding by leveraging pre-training over diverse data sets \cite{devlin2018bert}. While large language models (LLMs) like GPT, built on autoregressive transformer architectures, provide comprehensive language analysis via deep neural networks and attention mechanisms, they come with high computational demands. Their computational intensity and slower adaptation to novel, domain-specific terminology pose significant challenges.
They may not swiftly adapt to evolving domain-specific terms \cite{devlin2018bert}. Recognizing these constraints, we retain Word2Vec for its computational efficiency and ease of domain-specific fine-tuning. This approach ensures our model quickly adapts to the evolving linguistic landscape, like mental health, during the COVID-19 pandemic.

\subsection{Neurosymbolic AI}
Neurosymbolic AI harnesses the strengths of neural networks and symbolic AI, combining them to improve adaptability and effectiveness in processing dynamically evolving data such as social media content during health crises \cite{sheth2023neurosymbolic}. While neural networks excel in pattern recognition from vast data sets, they often lack transparency and can be data-intensive. Conversely, symbolic AI offers clear reasoning paths and requires less data but struggles with flexibility, which is critical for processing complex, unstructured datasets.

To address these challenges, our approach employs Knowledge-Infused Learning (K-iL), which strategically integrates structured knowledge bases at various levels within neural networks. This method effectively bridges the divide between data-driven and knowledge-driven processes, leveraging the strengths to enhance learning \cite{kursuncu2020knowledge,8970629}. KiL is implemented in three primary infusion levels, each designed to progressively incorporate deeper semantic understanding into the learning process, enhancing the model's ability to interpret and adapt to new information.

\textbf{Shallow Infusion:} This level involves minimal integration, often utilizing pre-trained embeddings to provide knowledge that enhances the neural network's understanding of domain-specific concepts without architectural changes \cite{8970629}.

\textbf{Semi-Deep Infusion:} At this intermediate level, knowledge is integrated into the learning process through focused attention mechanisms 
particularly enhancing their ability to interpret complex data structures \cite{8970629}.

\textbf{Deep Infusion:} The deepest level of knowledge infusion identifies where the latent weights are incorrectly applied within the model layers and adjusts these weights using an external, human-curated graphical knowledge source. \cite{8970629}. 
While deep infusion offers substantial advantages in accuracy and explainability, its implementation often relies on LLMs, which are computationally intensive. 

For applications that demand rapid adaptability and reduced computational resources, shallow and semi-deep infusions offer a solution. These techniques have been successfully applied and have shown improved performance across a range of NLP tasks, such as text entailment, classification, and question answering \cite{liu2020k, liu2020roberta, wang-etal-2021-k, he2020kgplm}. More details in Appendix~\ref{sec:appen_NeuroAI}.

\section{Methods}\label{sec:methods}

\begin{figure*}[ht]
    \centering
    \includegraphics[width=0.8\textwidth]{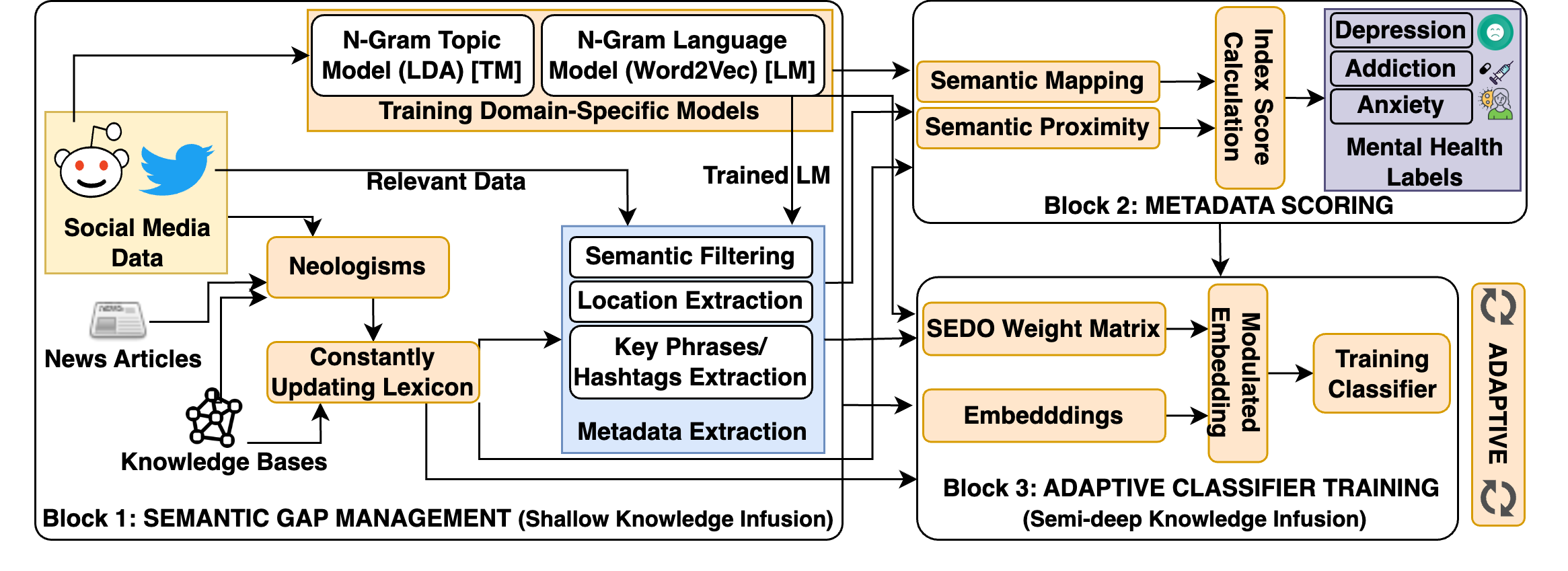}
    \caption{The architecture consists of three main components: B1 (Semantic Gap Management), B2 (Metadata Scoring), and B3 (Adaptive Classifier Training). In B1, employing shallow infusion, relevant data is gathered from diverse sources, including social media, news articles, and KBs. These processes include identifying new terms, updating lexicons, semantic filtering, location extraction, and key phrase/hashtag extraction. B2 involves semantic mapping and proximity for index score calculation, which is further used to obtain a classification label. In B3, featuring semi-deep knowledge infusion, classifiers are trained using extracted semantic data and metadata scores-based labels.}
    \label{fig:techapproach}
\end{figure*}

A unique aspect of our architecture (see Figure \ref{fig:techapproach}) is its methodological richness, which incorporates a diverse set of complementary data sources and multiple KBs. This setup facilitates the inclusion of new and relevant lexicons and utilizes two distinct knowledge infusion techniques tailored to meet different needs. Additionally, the approach supports a rich semantic evaluation through location extraction and index scoring, highlighting its novelty and practical utility.

The proposed method employs a multi-stage approach. The process begins with \textit{Semantic Gap Management} (\textit{B1}), where we utilize a diverse array of data sources, including social media data, news articles, and multiple KBs. We train domain-specific topic and language models, enriching them with contextual details such as location, key phrases, and updated lexicons. In \textit{Metadata Scoring} (\textit{B2}), semantic mapping and proximity are employed to label content relevant to the domain. The final stage, \textit{Adaptive Classifier Training} (\textit{B3}), involves semi-deep knowledge infusion techniques, including zero-shot training and fine-tuning of machine learning classifiers. These classifiers integrate domain-specific knowledge with extracted semantic data and metadata scores, enabling the architecture to adapt to dynamically evolving domains.

\subsection{Semantic Gap Management (B1)}
A "semantic gap" refers to the difference between the raw data available and the additional information that can be derived from it, transforming raw, unstructured data into structured, actionable insights. Semantic Gap Management, employing shallow infusion, involves broadly sourcing diverse raw data, including social media posts and news articles. Through preprocessing, we enhance the quality and relevance of this data; this includes tokenization, converting text to lowercase, removing high-frequency stopwords and punctuation, applying lemmatization, and generating n-grams.

The output of this phase is processed and enriched data featuring extracted and normalized n-gram key phrases, location information, domain-specific topics \& language models, and a dynamically updated lexicon. Our approach leverages multiple relevant KBs to  interpret the enriched data effectively
\cite{sheth2021knowledge}. This enriched data is structured and formatted, priming it for subsequent deeper semantic analysis and classification tasks, effectively bridging the semantic gap.

\subsubsection{Topic and Language Models}
We develop domain-specific models to analyze social media data related to health crises like the COVID-19 pandemic. 

\textbf{Topic Modeling:} We utilize LDA to identify latent topics from the text data. The process begins with constructing a vocabulary from our corpus, which is then transformed into a bag-of-words format.

\textbf{Language Modeling:} Concurrently, we train a language model to generate dense vector representations of n-grams that capture their semantic meanings. This model, while based on the principles of Word2Vec, is adapted to include contextual meanings by incorporating a range of context words specified by the Context Window Size. The training uses the Skip-gram approach, where the model predicts context words from target words and employs negative sampling \cite{mikolov2013distributed} to improve efficiency by focusing updates on the most relevant features.

Despite the potential computational demands of training models with multiple iterations, efficiency is maintained through negative sampling and incorporating KBs, which selectively update a subset of weights to speed convergence.

\subsubsection{Lexicons and Neologisms}
After training the Word2Vec model, we integrate specialized lexicons that serve as flattened knowledge bases. These lexicons feature a comprehensive list of keywords that cover a broad spectrum of terms, including slang, domain-specific terminology, and domain-expert vocabulary. Initially, these lexicons are developed by gathering a preliminary set of terms from existing domain-specific and gxeneral-purpose KBs, such as DAO and Dbpedia.

Given the dynamic nature of social media and evolving global events, it is crucial to continuously update our lexicon to maintain its relevance and accuracy in our analyses. We utilize data from various sources, including social media platforms and news articles. New terms are identified periodically from this collected data, effectively capturing emerging trends, novel contexts, and evolving language usage related to mental health topics. This ongoing process of lexicon enhancement ensures that we can adapt to shifting narratives, emerging themes, and the impacts of significant events or waves of events on conversations about mental health.

\subsubsection{Semantic Filtering}
Semantic filtering refines the relevancy of social media data for focused investigation of specific events and themes. This process is facilitated by a trained Word2Vec model, which generates embeddings for both the domain-specific lexicon and the social media content. 

To assess the relevance of social media posts, we calculate the cosine similarity between the lexicon's embeddings and those of the social media data. 
To establish a threshold for relevance, we analyzed the distribution of cosine similarity scores across a representative sample of the data. Based on this analysis, we selected the 75th percentile of the distribution as our threshold. Posts with cosine similarity scores exceeding this threshold (approximately 0.6) are deemed relevant and are subsequently included in the dataset for further analysis. This methodical approach ensures that our dataset retains content most pertinent to the investigated themes.

\subsubsection{Metadata Extraction}
Metadata extraction is crucial for analyzing mental health discussion distribution and regional variations. 
This process includes location and key phrase extraction, enriching the dataset's contextual understanding and relevancy to public health monitoring.

\textbf{Location Extraction:} We utilize the \textit{Geograpy} Python library to detect geographic information embedded within the metadata or the actual content of the data. This tool helps identify relevant location phrases cross-referenced with location knowledge bases to ascertain accurate high-level geographic details, such as state or region. 

\textbf{Key Phrase/Hashtag Extraction:} We employ a trained N-gram model to extract significant key phrases and hashtags pertinent to mental health topics from the social media data. Candidate phrases identified by the N-gram model are further scrutinized for relevance by measuring their semantic similarity to domain-specific lexicons. Only those phrases and hashtags that show high similarity are retained, thus ensuring that the analysis focuses on the most relevant content.

Our methods can be fine-tuned to other domains by continuously updating topic and language models, such as retraining LDA with n-grams, Word2Vec embeddings, and dynamically enriching lexicons with emerging terms.

\subsection{Metadata Scoring (B2)}
In the Metadata Scoring phase, the process leverages the enriched lexicon, key phrases, hashtags, and domain-specific topic and language models developed in the Content Enrichment stage to apply advanced semantic analysis. This analysis computes target labels for social media data, essential for the subsequent classifier training.

The enriched and preprocessed data from the earlier Semantic Gap Management phase (B1) serves as the input for this stage. Using this data, semantic mapping aligns relevant lexicon concepts with extracted key phrases and hashtags, enhancing the contextual representation of the data by connecting it with domain-specific knowledge. Simultaneously, semantic proximity identifies the closeness of topics to these key phrases and lexicons. Both semantic mapping and proximity use cosine similarity to compute these relationships.

Index scores, which serve as labels, are calculated based on the combined outcomes of semantic proximity and mapping as shown in Equation \ref{eq:index_score}:
\begin{equation}\label{eq:index_score}
H^S = \{ H(\text{ng}^S, D) + H(\text{LDA}^S, D) + H(\text{nLDA}^S, D) \}
\end{equation}
Here, \(H(\text{ng}^S, D)\) is the semantic mapping score between n-grams (\(ng^S\)) and the dataset (\(D\)), and \(H(\text{LDA}^S, D)\) along with \(H(\text{nLDA}^S, D)\) denote the semantic proximity scores using Latent Dirichlet Allocation (LDA) and n-gram LDA (nLDA), respectively. The normalization is performed as follows:
\begin{equation}
nhs_D^S = \frac{H_D^S}{\max(H^S)}
\end{equation}
This equation normalizes the semantic proximity score for a specific document or tweet (\(D\)) within the category (\(S\)) by dividing \(H_D^S\) by the maximum semantic score (\(\max(H^S)\)) within that category, ensuring that each category, such as Depression, Addiction, or Anxiety, has its index score scaled.

This calculated index score combines semantic proximity and mappings with n-grams and topics, effectively capturing the content's relevance to the lexicon concepts. The utility and accuracy of these computed labels, as benchmarks for training machine learning classifiers, will be assessed against human annotators in subsequent analyses (see Section \ref{subsec:validate_metascore}).

\subsection{Adaptive Classifier Training (B3)}

\begin{figure*}[ht]
    \centering
    \includegraphics[width=0.8\textwidth]{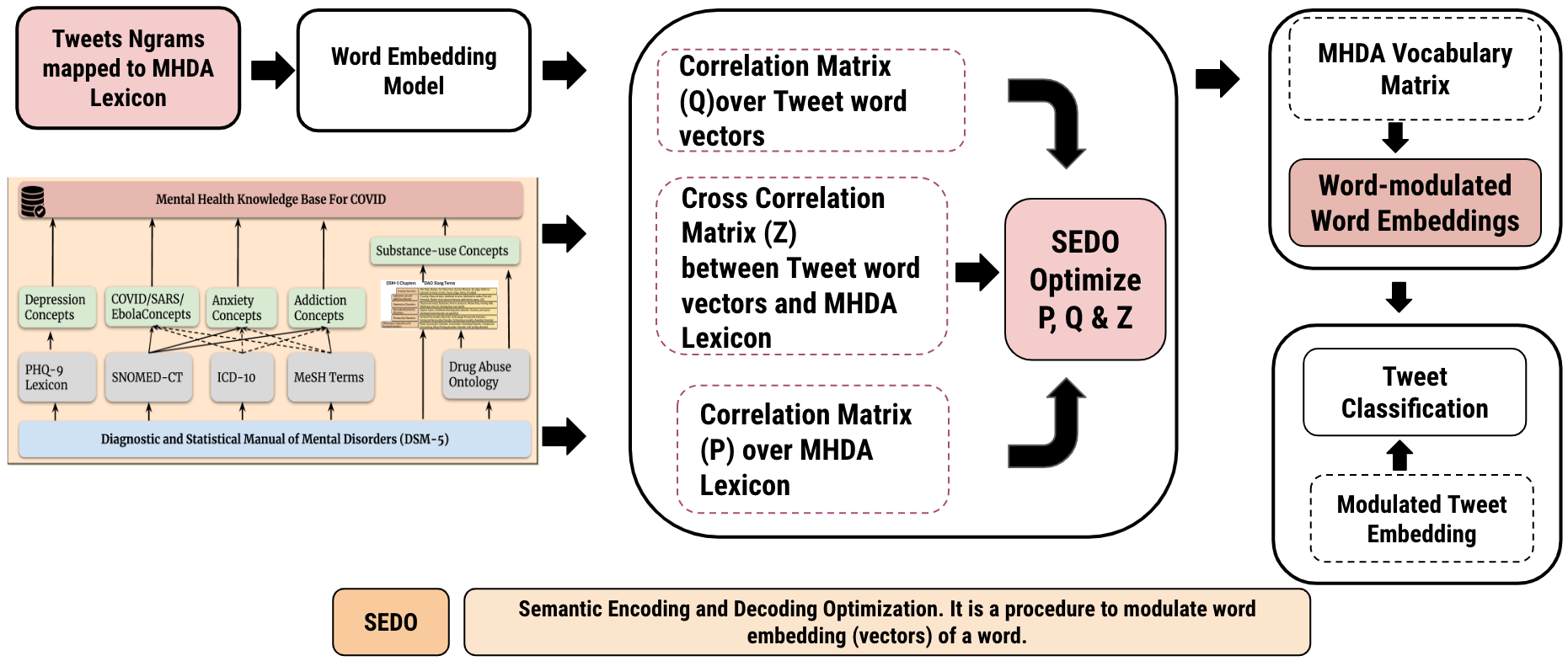}
    \caption{Overview of the SEDO framework for tweet classification related to mental health and addiction during COVID-19. The process begins with mapping tweet n-grams to the MHDA lexicon. The domain-specific language model generates initial word embeddings from these tweets. A correlation matrix \( Q \) is computed over the tweet word vectors, while a cross-correlation matrix \( Z \) is established between tweet word vectors and the MHDA lexicon. Additionally, a correlation matrix \( P \) is derived over the MHDA lexicon. The SEDO procedure optimizes the matrices \( P \), \( Q \), and \( Z \), modulating the word embeddings through a learned weight matrix. The modulated word embeddings are then utilized for tweet classification. The MHDA KBx integrates concepts from domain-specific resources, including the  PHQ-9 lexicon, SNOMED-CT, ICD-10, MeSH terms, and the DAO, enriching the analysis with domain-specific vocabulary and relationships.}
    \label{fig:sedo_framework}
\end{figure*}

The final module in our analysis framework involves training machine learning classifiers, 
using the labeled data generated by the Metadata Scoring module. This stage, employing semi-deep knowledge infusion, utilizes the SEDO method, which incorporates the Sylvester equation \cite{sylvestereqtopicmodel2010} to develop a discriminative weight matrix. This matrix modulates the word embeddings of tweets based on their semantic proximity to mental health and drug abuse (MHDA) lexicons. By dynamically adjusting embeddings to reflect evolving terminology and context, this method enhances the classifiers' ability to generalize and maintain accuracy across diverse social media datasets. Figure \ref{fig:sedo_framework} provides an illustrative diagram of the SEDO framework.

The input for this phase consists of labeled data that indicate the presence or absence of specific mental health and drug abuse content, forming the basis for binary classification. The output is a set of robustly trained classifiers, each optimized to accurately identify a specific category of mental health-related content as separate binary classification tasks.

\subsection{Semantic Encoding and Decoding Optimization (SEDO)}
We employed the SEDO approach to optimize the integration of domain-specific knowledge into word embeddings. In this approach, \( T \) is the Twitter word embedding space, \( M \) is the embedding space of the mental health and drug abuse knowledge base (MHDA-Kb), and \( W \) is the learned weight matrix. The regularization parameter \( \delta \) controls the balance between alignment and regularization. The optimization is based on Sylvester equation:

\begin{equation}
(MM^T) W + W (\delta TT^T) = (1 + \delta) MT^T
\end{equation}

This equation finds the optimal weight matrix \( W \) that aligns the Twitter and MHDA-Kb embedding spaces, allowing the model to adapt dynamically to evolving terminologies and improve classification accuracy. More details in Appendix~\ref{sec:appen_sedo}.

\section{Experimental Setup}\label{sec:experimentalsetup}

\subsection{Resources}\label{subsec:resources}
We utilize 
social media datasets, news articles,
domain-specific and extensive general-purpose KBs.

\subsubsection{Twitter Dataset}\label{subsec:twitter}
Our Twitter dataset covers the period from March 14, 2020, through January 31, 2021, capturing the onset and two significant waves of the COVID-19 pandemic. This period is critical for analyzing the evolving public sentiment during notable instability. We collected approximately 12 billion tweets using 77 distinct keywords, including specific medical terms like ``covid-19,'' ``sars,'' ``lung,'' ``fever,'' and medications such as ``ritonavir,'' as well as more general terms like ``decreased\_white\_blood\_cells.'' Sample tweets include: ``Just tested positive for COVID-19, feeling scared \#fever \#cough'' and ``Wearing masks to stop the spread of SARS \#publichealth.'' This dataset provides a comprehensive snapshot of public discourse related to the pandemic and mental health. A major challenge with the raw data includes noise, irrelevant information, and missing location metadata. 

\subsubsection{Reddit Dataset}\label{subsec:reddit}
To complement our Twitter analysis, we also utilized data from Reddit, a platform structured into topic-specific forums known as ``subreddits" with posts that are less space-limited (extended to 40,000 characters) and, therefore, capture more explicit context. 
We analyzed 2.5 million posts (including main posts, replies, and comments) by 268,000 users from subreddits related to Drug Abuse, Opiates, Addiction, Anxiety, and Depression from 2005 to 2016 and the "coronavirus" subreddit from 2019 to 2020 \cite{gaur2018let}. 
Sample posts include: "Struggling with depression during these hard times. Anyone else feeling the same? \#MentalHealth" and "How do you guys manage anxiety with all the COVID news?"

\subsubsection{News Articles}\label{subsec:news}
News articles 
help us understand societal impacts, policy developments, and emerging mental health trends. We collected 700,000 COVID-19-related news articles from January 1 to March 29, 2020, using web-crawling APIs. Examples of such articles include headlines like ``Global pandemic: COVID-19 impacts public health and economy" and ``Mental health toll rises as pandemic continues."

\subsubsection{Knowledge Bases}\label{subsec:knowledge_bases}
We introduce and utilize the Mental Health and Drug Abuse Knowledge Bases (MHDA-Kb) alongside general-purpose KBs.

\noindent\textbf{Mental Heath and Drug Abuse-Knowledge Base (MHDA-Kb):} The MHDA-Kb encompasses specialized resources tailored to the unique terminologies and diagnostic criteria used in the mental health field. This includes:
\begin{itemize}
    \item \textbf{DSM-5:} Diagnostic criteria for mental disorders \cite{guha2014diagnostic}. Provides diagnostic criteria for mental disorders, essential for identifying mental health conditions.
    \item \textbf{DAO:} Contains terms related to drugs, mental health conditions, and their slang as used on social platforms \cite{lokala2022drug}.
    \item \textbf{SNOMED-CT:} Standardized clinical terms \cite{donnelly2006snomed}. It offers comprehensive clinical terminology for accurately coding and sharing clinical data.
    \item \textbf{UMLS:} Unified medical terminologies \cite{lindberg1990unified}. Integrates diverse health and biomedical terminologies, enhancing data interoperability across systems.
    \item \textbf{PHQ-9 Lexicon:} Terms from the depression assessment tool \cite{kroenke2001phq}. Includes terms related to depression severity assessments, useful for mental health evaluations.
\end{itemize}

\noindent\textbf{General-Purpose Knowledge Bases:}
\begin{itemize}
    \item \textbf{DBpedia:} Structured content from Wikimedia projects \cite{lehmann2015dbpedia}, aids in identifying and disambiguating entities.
    \item \textbf{WikiData:} Open knowledge base for Wikipedia's data \cite{vrandevcic2014wikidata}. Provides an editable knowledge base linked to Wikipedia's structured data for enhanced understanding.
    \item \textbf{ConceptNet:} Semantic network for understanding word relationships \cite{speer2017conceptnet}. A semantic network that represents word meanings and relationships across languages.
    \item \textbf{GeoNames and OpenStreetMap:} Tools for location extraction and mapping \cite{map2017open}. Provide geographic information and mapping capabilities to pinpoint locations mentioned in social media posts.
\end{itemize}

\subsection{Data Preparation and Preprocessing}
Our initial dataset of approximately 12 billion tweets was 
filtered as indicated in Sec.~\ref{subsec:resources} to capture tweets containing location information, resulting in 12 billion tweets. Further refinement was made by removing irrelevant "noisy" tweets based on semantic similarity with the relevant COVID-19 lexicon, which narrowed the dataset to 900 million tweets. As indicated in  Sec.~\ref{subsec:resources} More targeted filtering using MHDA lexicons yielded 600 million tweets. We also utilized 2.5 million posts from Reddit, training domain-specific Word2Vec and LDA topic models. These models also processed key phrases from news articles and hashtags from tweets, enhancing our lexicons with neologisms and emergent terms related to the pandemic's evolving context.

\noindent\textbf{Neologisms:} 
During the pandemic, we observed 
the emergence of specific terms across different platforms. For example, hashtags like \#trumppandemic and \#coronapocalypse on Twitter mirrored public sentiment and reactions. At the same time, Reddit discussions brought forth terms like \textit{PandemicBrain} and \textit{ZoomFatigue}, highlighting personal struggles and adaptations. News media also adapted language, with terms like \textit{economic\_instability} becoming prevalent, reflecting broader societal concerns. More details in Appendix~\ref{sec:appen_neologism}.

\subsection{Validation of Metadata Scoring}\label{subsec:validate_metascore}

We conducted a human annotation study with a subset of 500 data points each for depression, addiction, and anxiety. Three independent annotators reviewed the data, achieving a Cohen’s Kappa score of 0.72, which indicates substantial agreement among them. This consolidated label, formed using majority rule, served as a reference standard. When comparing our model’s labels with this standard, the confirmed accuracy rate was consistent with the inter-annotator agreement.

\subsection{Classifier Training and Evaluation}
We utilized an 80-20 split across different time intervals for classifier training to optimize our classifier parameters, including the SEDO weight matrix. This matrix was updated regularly with weights for approximately 30,000 phrases, facilitating the modulation of word embeddings used in training. We employed a range of classifiers: Naive Bayes, Random Forest, and variations of Balanced Random Forest—tailored to handle imbalanced data, which is paramount for identifying relevant content in vast social media datasets. Each mental health category—depression, addiction, and anxiety—was treated as a separate binary classification problem. This approach allows for more focused model training and accuracy, classifying each tweet as relevant or irrelevant to each category.

\noindent\textbf{Triangulation Study: }To validate the generalizability and robustness of our classifiers, we conducted a triangulation study using published social media datasets manually annotated for depression, addiction, and anxiety. This approach helps ensure our models are effective across different data sources and real-world scenarios. By testing our model on unrelated, published social media datasets for depression \cite{nadeem2016identifying}, addiction \cite{glowacki2021identifying}, and anxiety \cite{owen2020towards}, we aimed to demonstrate its effectiveness and reliability across different data sources. Below are the two experiments performed on these datasets:

\begin{itemize}
    \item \textbf{Experiment 1: Pre-trained SEDO Weight Matrix}
We tested the model using the pre-trained SEDO weight matrix in the first experiment. This setup allowed us to evaluate how well the model, trained on our original dataset, could generalize to the new validation dataset without any additional fine-tuning.

    \item \textbf{Experiment 2: Fine-tuned SEDO Weight Matrix}
We fine-tuned the SEDO weight matrix on the new validation dataset in the second experiment. This approach aimed to assess the improvement in classification performance when the model's parameters were adapted to the specific characteristics of the new data.

\end{itemize}

\noindent\textbf{Ablation Study: } This study evaluates the relevance and impact of several components of our analysis. Specifically, we measure component relevance based on the change in error rates. A qualitative approach showcases how components contribute to the overall analysis. We use "green" and "red" examples to illustrate successes and limitations, respectively, in our model's performance:
 
\begin{itemize}
    \item \textbf{Green Examples:} These examples highlight instances where our model successfully identifies and classifies content accurately, demonstrating the effectiveness of integrating specific resources or techniques.
    \item \textbf{Red Examples:} These examples reveal situations where the model version fails to correctly interpret and classify the data, pointing to areas where improvements are needed or showcasing the challenges inherent in processing complex social media content.
\end{itemize}

We conducted the above-mentioned experiments to evaluate the performance of our proposed framework. Subsequent sections detail the results and insights from the analysis.

\noindent\textbf{Comparison with Large Language Models (LLMs)}
We also compare the performance of our models against state-of-the-art large language models (LLMs) to highlight our models' computational efficiency and effectiveness. We curated a dataset comprising 1,000 tweets per category (depression, addiction, and anxiety) from three different time frames: April-May 2020, August-September 2020, and December 2020 - January 2021. This temporal segmentation allowed us to assess the models' ability to adapt to new and evolving terminology across different periods. The LLMs—LLama (7 billion parameters) \cite{touvron2023llama}, Phi (2.7 billion parameters) \cite{javaheripi2023phi}, and Mistral (7 billion parameters) \cite{jiang2023mistral}—were evaluated using their open-source instruct-tuned versions in a zero-shot setting. This approach ensures an apple-to-apple comparison with our classifiers, which also process only the tweet converted into embeddings without additional context.

\section{Results}

This section summarizes the results of our experiments, emphasizing the effectiveness of our data preparation and classification methods and the impact of ablation and triangulation studies. We utilize Naive Bayes (NB), Random Forest (RF), Balanced Random Forest(BRF), and Balanced Sub-Sample Random Forest(BSRF) models for the experiment. 

\noindent \textit{Overall:} The application of the SEDO weight matrix substantially enhanced the performance metrics of the models, as evidenced by the data presented in Table~\ref{tab:classifier_performance}. The incorporation of the SEDO matrix resulted in improvements in precision, recall, and F1-Score across various classifiers. The values in red parentheses in the table denote the percentage decrease in performance metrics when the SEDO matrix is excluded, underscoring its significant role in boosting the effectiveness of the classifiers across all evaluated metrics.

\begin{table}[ht]
\centering
\caption{Classifier performance with SEDO and without SEDO (\% decrease values in red color).}
\label{tab:classifier_performance}
\begin{tabular}{|p{1.2cm}|p{0.8cm}|l|l|l|}
\hline
\textbf{Category}                             &\textbf{Model}           & \textbf{Precision}      & \textbf{Recall}         & \textbf{F1-Score}       \\ \hline
\multirow{4}{*}{\textbf{Depression}}       & NB             & 84.85 (\textcolor{red}{-24\%})           & 82.68 (\textcolor{red}{-25\%})           & 83.75 (\textcolor{red}{-27\%})           \\ \cline{2-5} 
                                           & RF & 91.98 (\textcolor{red}{-28\%})           & 91.81 (\textcolor{red}{-26\%})           & 91.89 (\textcolor{red}{-23\%})           \\ \cline{2-5} 
                                           & BRF   & 92.32 (\textcolor{red}{-27\%})           & 92.43 (\textcolor{red}{-24\%})           & 92.37 (\textcolor{red}{-29\%})           \\ \cline{2-5} 
                                           & BSRF   & \textbf{94.12} (\textcolor{red}{-29\%})           & \textbf{93.02} (\textcolor{red}{-22\%})           & \textbf{93.57} (\textcolor{red}{-28\%})           \\ \hline\hline
\multirow{4}{*}{\textbf{Addiction}}        & NB             & 82.74 (\textcolor{red}{-26\%})           & 80.46 (\textcolor{red}{-21\%})           & 81.58 (\textcolor{red}{-25\%})           \\ \cline{2-5} 
                                           & RF & 90.02 (\textcolor{red}{-22\%})           & 90.36 (\textcolor{red}{-20\%})           & 90.19 (\textcolor{red}{-23\%})           \\ \cline{2-5} 
                                           & BRF   & 91.53 (\textcolor{red}{-28\%})           & 91.78 (\textcolor{red}{-26\%})           & \textbf{91.65} (\textcolor{red}{-29\%})           \\ \cline{2-5} 
                                           & BSRF   & \textbf{91.64} (\textcolor{red}{-27\%})           & \textbf{91.82} (\textcolor{red}{-24\%})           & 91.73 (\textcolor{red}{-28\%})           \\ \hline\hline
\multirow{4}{*}{\textbf{Anxiety}}          & NB             & 82.53 (\textcolor{red}{-25\%})           & 81.87 (\textcolor{red}{-24\%})           & 82.20 (\textcolor{red}{-22\%})           \\ \cline{2-5} 
                                           & RF & 90.76 (\textcolor{red}{-23\%})           & 92.78 (\textcolor{red}{-28\%})           & 91.76 (\textcolor{red}{-21\%})           \\ \cline{2-5} 
                                           & BRF   & \textbf{94.37} (\textcolor{red}{-27\%})           & \textbf{93.87} (\textcolor{red}{-25\%})           & \textbf{94.12} (\textcolor{red}{-29\%})           \\ \cline{2-5} 
                                           & BSRF   & 93.46 (\textcolor{red}{-24\%})           & 93.85 (\textcolor{red}{-27\%})           & 93.65 (\textcolor{red}{-28\%})           \\ \hline
\end{tabular}
\end{table}

\subsection{Triangulation Study}

To assess the generalizability and adaptability of our SEDO framework, we conducted a triangulation study comprising two experiments: the first using the SEDO matrix trained earlier on the COVID-19 social media data, and the second with SEDO fine-tuned on unrelated, published social media datasets addressing depression \cite{nadeem2016identifying}, addiction \cite{glowacki2021identifying}, and anxiety \cite{owen2020towards}. These experiments tested our framework's 
robustness and enhancement capacity through fine-tuning on unseen data.

\begin{table}[ht]
\centering
\caption{Experiment 1: Classifier performance with SEDO.}
\label{tab:exp1_performance}
\begin{tabular}{|p{1.2cm}|p{0.8cm}|l|l|l|}
\hline
\textbf{Category}  & \textbf{Model}                             & \textbf{Precision}      & \textbf{Recall}         & \textbf{F1-Score}       \\ \hline
\multirow{4}{*}{\textbf{Depression}}       & NB             & 75.43                   & 74.64                   & 75.03                   \\ \cline{2-5} 
                                           & RF & 84.32                   & 81.42                   & 82.84                   \\ \cline{2-5} 
                                           & BRF   & 83.56                   & \textbf{84.29}                   & 83.92                   \\ \cline{2-5} 
                                           & BSRF   & \textbf{85.65}                   & 84.14                   & \textbf{84.89}                   \\ \hline \hline
\multirow{4}{*}{\textbf{Addiction}}        & NB             & 74.28                   & 74.27                   & 74.28                   \\ \cline{2-5} 
                                           & RF & 80.66                   & 79.22                   & 79.93                   \\ \cline{2-5} 
                                           & BRF   & 80.94                   & 82.06                   & 81.49                   \\ \cline{2-5} 
                                           & BSRF   & \textbf{83.27}                   & \textbf{82.49}                   & \textbf{82.88}                   \\ \hline \hline
\multirow{4}{*}{\textbf{Anxiety}}          & NB             & 72.68                   & 73.21                   & 72.94                   \\ \cline{2-5} 
                                           & RF & 79.46                   & 82.35                   & 80.88                   \\ \cline{2-5} 
                                           & BRF   & \textbf{83.72}                   & \textbf{82.49}                   & \textbf{83.10}                   \\ \cline{2-5} 
                                           & BSRF   & 83.47                   & 81.63                   & 82.54                   \\ \hline
\end{tabular}
\end{table}

The results from Experiment 1 (as shown in Tab.~\ref{tab:exp1_performance}, using the SEDO matrix trained on COVID-19 data, demonstrate robust classifier performance across all categories, substantiating the applicability of our SEDO framework to unseen datasets. Notably, the BSRF model excelled in the Depression category, with an F1-Score of 84.89\%, highlighting model’s robustness.

\begin{table}[ht]
\centering
\caption{Experiment 2: Classifier performance with fine-tuned SEDO.}
\label{tab:exp2_performance}
\begin{tabular}{|p{1.2cm}|p{0.8cm}|l|l|l|}
\hline
\textbf{Category}  & \textbf{Model}                             & \textbf{Precision}      & \textbf{Recall}         & \textbf{F1-Score}       \\ \hline
\multirow{4}{*}{\textbf{Depression}}       & NB             & 78.90                   & 80.82                   & 79.85                   \\ \cline{2-5} 
                                           & RF & 84.37                   & 86.16                   & 85.26                   \\ \cline{2-5} 
                                           & BRF   & \textbf{89.42}                   & 89.79                   & \textbf{89.60}                   \\ \cline{2-5} 
                                           & BSRF   & 87.65                   & \textbf{90.72}                   & 89.16                   \\ \hline \hline
\multirow{4}{*}{\textbf{Addiction}}        & NB             & 79.90                   & 81.63                   & 80.76                   \\ \cline{2-5} 
                                           & RF & 85.83                   & 85.52                   & 85.68                   \\ \cline{2-5} 
                                           & BRF   & \textbf{89.39}                   & \textbf{89.87}                   & \textbf{89.63}                   \\ \cline{2-5} 
                                           & BSRF   & 88.84                   & 89.38                   & 89.11                   \\ \hline \hline
\multirow{4}{*}{\textbf{Anxiety}}          & NB             & 80.75                   & 79.64                   & 80.19                   \\ \cline{2-5} 
                                           & RF & 87.81                   & 86.78                   & 87.29                   \\ \cline{2-5} 
                                           & BRF   & 87.98                   & 87.92                   & 87.95                   \\ \cline{2-5} 
                                           & BSRF   & \textbf{89.14}                   & \textbf{90.59}                   & \textbf{89.86}                   \\ \hline
\end{tabular}
\end{table}

Experiment 2 demonstrated the benefits of fine-tuning SEDO,  improving classifier metrics. Here, fine-tuning refers to updating the SEDO weights based on the new phrases and their relevance with depression, addiction and anxiety. The BSRF model achieved an F1-Score of 89.16\% for the depression category (a 5.03\% increase compared to BSRF + SEDO trained on COVID-19 data), underscoring our framework's adaptability to optimize performance through fine-tuning. This suggests that while the initial SEDO is highly effective, its performance can be further improved with fine-tuning on unseen data.

\subsection{Ablation Study}
Our ablation study highlights the contribution of general-purpose KBs, domain-specific KBs, and pre-trained and fine-tuned language models within our analysis framework. The inclusion of each component is shown to reduce error rates, progressively enhancing classification accuracy. General-purpose KBs expand the contextual understanding of data, improving the model's applicability across varied discussions (as shown in Tab.~\ref{tab:example_classification}). The subsequent addition of domain-specific KBs tailors this approach, sharpening the focus on mental health and addiction-related content, further improving accuracy.

Further refinement through domain-specific fine-tuning of language models demonstrates marked improvements in handling specialized content, significantly reducing error rates. Detailed qualitative and quantitative analyses in Appendix~\ref{sec:appen_ablation} illustrate these improvements, showcasing each component's indispensable role in improving our analytical framework's adaptability and precision.

\subsection{Comparison with LLMs}

To highlight the effectiveness of our neurosymbolic approach 
, we compared its performance against state-of-the-art LLMs like LLama, Phi, and Mistral. As shown in Table \ref{tab:llm_comparison}, our approach consistently outperformed the LLMs in precision, recall, and F1-score across the Depression, Addiction, and Anxiety categories, with values between 89-93.6\%, while the LLMs ranged between 70-80\%.\begin{table}[ht]
    \centering
    \caption{Comparison with LLMs against the Neurosymbolic Approach.}
    \begin{tabular}{|l|c|c|c|c|}
        \hline
        \textbf{Category} & \textbf{Model} & \textbf{Precision} & \textbf{Recall} & \textbf{F1-Score} \\
        \hline
        \multirow{4}{*}{Depression} & LLama & 74.23 & 70.57 & 72.34 \\
        \cline{2-5}
        & Phi & 71.67 & 66.42 & 68.95 \\
        \cline{2-5}
        & Mistral & 76.51 & 71.38 & 73.87 \\
        \cline{2-5}
        & \textbf{Neurosymbolic} & \textbf{90.45} & \textbf{87.29} & \textbf{88.84} \\
        \hline
        \multirow{4}{*}{Addiction} & LLama & 77.24 & 73.68 & 75.42 \\
        \cline{2-5}
        & Phi & 73.32 & 69.75 & 71.49 \\
        \cline{2-5}
        & Mistral & 78.45 & 74.67 & 76.51 \\
        \cline{2-5}
        & \textbf{Neurosymbolic} & \textbf{92.18} & \textbf{88.36} & \textbf{90.22} \\
        \hline
        \multirow{4}{*}{Anxiety} & LLama & 78.56 & 74.82 & 76.66 \\
        \cline{2-5}
        & Phi & 74.38 & 70.61 & 72.43 \\
        \cline{2-5}
        & Mistral & 80.33 & 76.89 & 78.56 \\
        \cline{2-5}
        & \textbf{Neurosymbolic} & \textbf{93.25} & \textbf{90.52} & \textbf{91.85} \\
        \hline
    \end{tabular}
    \label{tab:llm_comparison}
\end{table}

\noindent\textbf{Efficiency:} In addition to superior accuracy, our neurosymbolic model achieved faster convergence times than LLMs. The RF converged in about 40 minutes, while the BSRF took 55 minutes. In contrast, a typical LLM requires approximately $>6-8$ hours to converge under similar conditions ($>500k$ training examples). This demonstrates a significant efficiency improvement in our approach regarding training time.


\subsection{Error Analysis}

 LLMs such as LLama and Phi struggle with emerging slang and jargon, such as ``doomscrolling" or ``quarantini," which surfaced during the pandemic. These terms, absent from their training datasets, often lead to misclassifications, highlighting LLMs' challenges with rapidly evolving language.

Our SEDO-enhanced classifiers, designed to be more adaptable through knowledge infusion, still encountered significant hurdles. For example, a tweet using the phrase ``I'm totally gutted," common in British English to express profound disappointment, was incorrectly interpreted by SEDO as expressing physical discomfort due to its reliance on primarily American English sources. In another instance, the phrase ``catching these hands" is common in American slang to indicate an impending fight, yet it was misinterpreted literally by our system as physically catching hands, missing the metaphorical violence implied. Additionally, a tweet containing the Australian slang ``I'm devo," shorthand for ``devastated," was inaccurately processed as having a neutral sentiment due to the system's unfamiliarity with this abbreviation. These challenges are not exclusive to SEDO-enhanced models; large language models (LLMs) such as LLama and Phi face similar issues.

\section{Discussion}

In this study, we have developed a multi-stage neurosymbolic framework for analyzing mental health discussions on social media, specifically Twitter, during the COVID-19 pandemic. Our approach utilizes domain-specific KBs such as DAO, and UMLS, improving the accuracy of content classification. Utilizing shallow and semi-deep knowledge infusion techniques allows for robust analysis of large-scale social media data, moving beyond the constraints of simple keyword-based approaches.

Our framework integrates domain-specific and general-purpose knowledge. This strategic incorporation is crucial for extracting relevant content from unstructured big social media data, addressing a significant challenge in big data analysis. The system's capability to identify and incorporate emerging terms from Twitter, Reddit, and News Articles ensures that our lexicons evolve with online discourse, maintaining the relevance and accuracy of the analysis amidst evolving contexts.

Moreover, the neurosymbolic approach optimizes the trade-off between accuracy and efficiency, which is often challenging with traditional methods. By incorporating the SEDO adaptation and utilizing less computationally intensive models like Word2Vec, our framework achieves quick updates in response to dynamic events. This adaptive strategy allows for practical real-time applications, effectively managing the complexity and scale of big data. It ensures that our analysis stays current with the rapidly evolving landscape of social media discourse, making it particularly suited to real-time applications where timely data processing is critical. This ongoing development demonstrates the practical benefits of integrating neurosymbolic methods with machine learning classifiers, promising valuable insights into public health and other areas of societal importance.

\section{Ethics and Reproducibility}
Our study has been granted an IRB waiver from the University of South Carolina IRB (application number Pro00094464). To further support ethics and reproducibility, we share our code and dataset via GitHub, accessible at \url{https://tinyurl.com/nesy-cv19}.

\section{Conclusion and Limitations}
We introduced a neurosymbolic approach to analyze Twitter data for mental health during the COVID-19 pandemic. By integrating neural networks with symbolic knowledge sources, our method overcame the limitations of purely data-driven models, which often fail to capture context due to the ever-changing nature of social media language. This combination of general-purpose and domain-specific KBs enabled our model to achieve an F1 score above 92\%. Despite this, our approach cannot respond to regional and cultural slang terms. This limitation highlights the need to incorporate a wider variety of linguistic inputs and update knowledge bases dynamically to handle culturally specific expressions and regional slang better, ensuring more accurate and culturally sensitive classifications.

Future research can focus on mapping mental health trends identified in our study to policy decisions, exploring how public sentiment responds to new health interventions. By correlating these trends with policy changes, we aim to provide valuable insights for policymakers during crises. Leveraging neurosymbolic AI will enhance public health monitoring, enabling more informed and timely interventions.
Finally, the data collection period was confined to 2020-2021, and the evolving nature of social media platforms, particularly X, presents challenges for future research. Replicating the scale of this data collection under X's updated API policies would be more difficult and costly. To address this, we have shared the X IDs of the collected tweets, allowing other researchers to access and analyze the dataset within the current API constraints.



\bibliographystyle{IEEEtrans}
\bibliography{IEEEabrv, IEEEexample, ref}

\appendices

\section{Neurosymbolic AI}\label{sec:appen_NeuroAI}

Integrating symbolic knowledge into neural network models, known as neurosymbolic AI, marks a significant advancement in NLP. While adept at detecting latent patterns, traditional deep learning models often struggle with context and semantics, particularly in specialized domains. By incorporating structured knowledge from Knowledge Bases (KBs), neurosymbolic models gain additional context and relational insights, overcoming the limitations of purely data-driven approaches \cite{kursuncu2020knowledge, bian2014knowledge}.


One prominent approach is K-BERT, which integrates knowledge graphs into the BERT model to enhance language representation capabilities \cite{liu2020k}. By injecting domain-specific knowledge from KGs into sentences, K-BERT constructs knowledge-rich sentence trees, which help maintain the original sentences' correct meaning while introducing relevant context. This method has demonstrated superior performance in domain-specific tasks, such as finance, law, and medicine, compared to the vanilla BERT model.

Another notable approach is K-ADAPTER, which leverages compact neural models known as adapters. These adapters integrate factual and linguistic knowledge into pre-trained models like RoBERTa, significantly enhancing performance across multiple knowledge-driven tasks, including relation classification, entity typing, and question answering \cite{wang-etal-2021-k}. This method effectively captures versatile knowledge and achieves state-of-the-art results on several benchmark datasets.

The concept of Knowledge-Infused Learning (K-IL) further explores the deep integration of knowledge within neural networks \cite{8970629}. This approach emphasizes the importance of incorporating domain knowledge directly within the latent layers of neural models. By doing so, it aims to address challenges such as sparse feature occurrence, ambiguity, and noise in data. K-IL frameworks have shown improvements in tasks requiring deep semantic understanding and contextual awareness by modulating word vectors with knowledge from KGs during the representation learning phase \cite{rawte2022tdlr}.

Furthermore, the work on KGPLM introduces a pre-training method that simultaneously models generative and discriminative knowledge approaches \cite{he2020kgplm}. This model achieves state-of-the-art performance on several question-answering datasets by effectively incorporating structured knowledge during the pre-training phase, demonstrating the efficacy of knowledge-guided learning for complex NLP tasks.

These advancements underscore the critical role of external knowledge in enhancing the performance and interpretability of neural network models. Integrating structured knowledge improves the accuracy of these models in specific domains. It provides a mechanism for robustness, making them more reliable and effective for practical applications.

\section{Neologisms}\label{sec:appen_neologism}
Our method continuously updates its enriched lexicon to include new terms from social media platforms, news articles, and domain-specific knowledge graphs, adapting to the dynamic nature of global events and social media. This ongoing process allows our models to remain relevant and accurate by integrating emerging trends and evolving language usage related to mental health.

We first map content from various sources to our lexicons to assess the prevalence and context of mental health terms, identifying both commonalities and unique insights each platform offers. We then analyze how language and sentiment evolve, uncovering shifts in narrative and emerging themes, particularly in response to significant events like the COVID-19 pandemic stages.

\begin{figure}[!htbp]
\centerline{\includegraphics[width=0.98\linewidth]{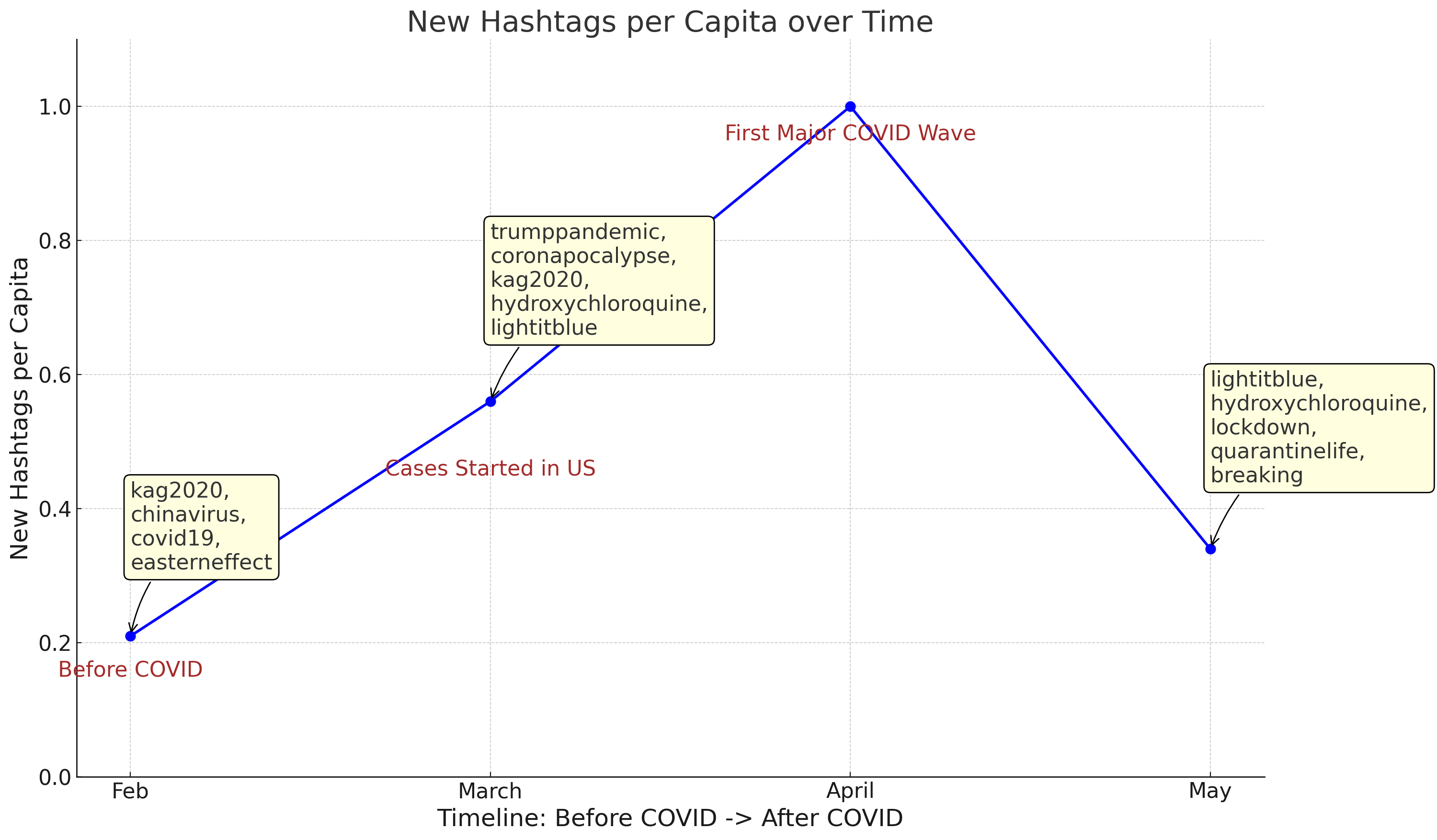}}
\caption{New terms from Twitter data}
\label{fig:newtwitter}
\end{figure}

\begin{figure}[!htbp]
\centerline{\includegraphics[width=0.98\linewidth]{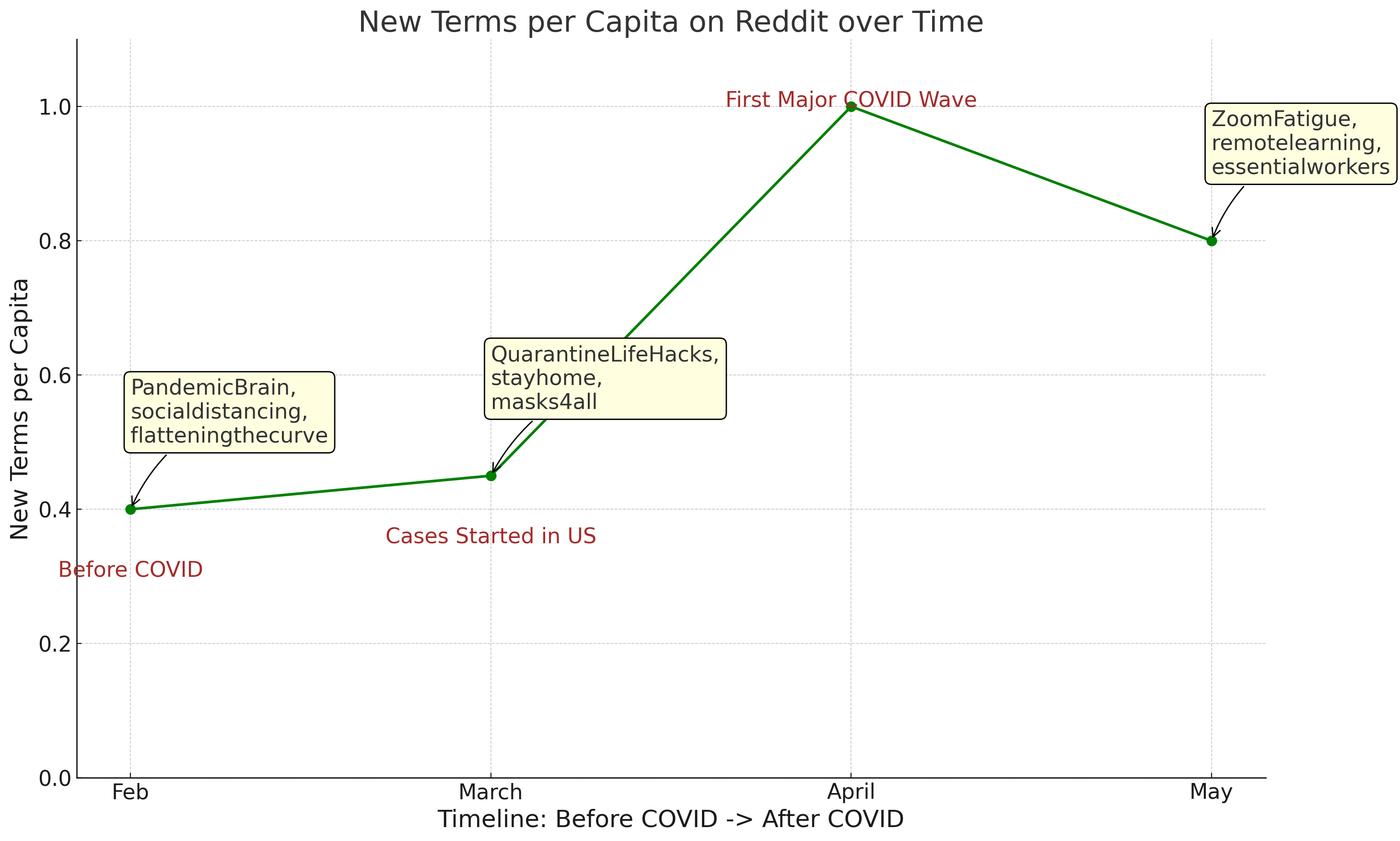}}
\caption{New terms from Reddit data}
\label{fig:newreddit}
\end{figure}

The enriched dataset includes location data, key phrases filtered for semantic relevance, and other pertinent information. For instance, from Twitter, terms such as \#trumppandemic and \#coronapocalypse emerged, captured in Fig.~\ref{fig:newtwitter}, illustrating hashtags' frequency and context during different pandemic stages. Similarly, Reddit discussions brought forward terms like \textit{PandemicBrain} and \textit{ZoomFatigue}, shown in Fig.~\ref{fig:newreddit}, offering insights into the public's pandemic experiences. News data, analyzed through web-crawling and entity extraction via ConceptNet and DBpedia, highlighted terms like \textit{American\_virus}, providing a snapshot of the socio-economic impacts and narrative shifts (Fig.~\ref{fig:newreddit}).

\begin{figure}[!htbp]
\centerline{\includegraphics[width=0.98\linewidth]{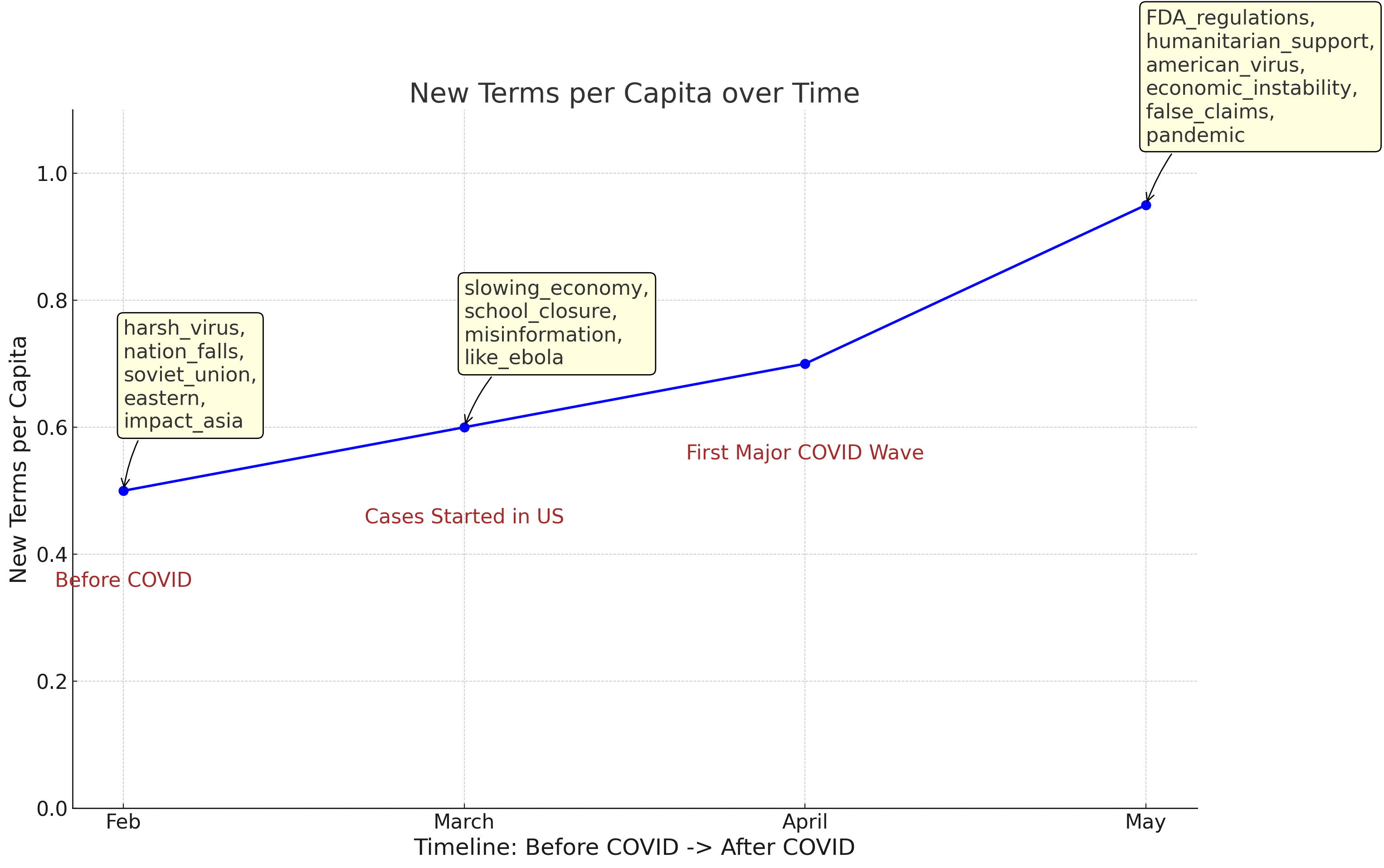}}
\caption{New terms from News articles}
\label{fig:newnews}
\end{figure}

In the figures, "per capita" refers to the ratio of the frequency of each term to the total mentions of all terms within the dataset, normalized per 1000 posts. This normalization helps standardize data representation despite variations in sample size or user activity across different platforms or time periods, allowing for more consistent comparisons of term prevalence.

\section{Semantic Encoding and Decoding Optimization (SEDO)} \label{sec:appen_sedo}

The SEDO approach provides
a form of semi-deep knowledge infusion. By modulating word embeddings through a learned weight matrix, SEDO enables our model to adapt to evolving terminologies and contextual meanings without extensive retraining. This method integrates domain-specific knowledge bases directly into the embedding process.
SEDO's capability to dynamically adjust embeddings aligns with the principles of semi-deep infusion, where external knowledge is actively incorporated into the learning process to improve the model's representational power and predictive accuracy.

The SEDO approach can be mathematically framed using the Sylvester equation. This equation is central to an encoding-decoding process, where we aim to optimize the transformation between the Twitter embedding space and the DSM-5-enriched mental health and drug abuse knowledge base lexicon (MHDA-Kb).

Let \( T \) represent the Twitter word embedding space and \( M \) the MHDA-Kb embedding space. The goal is to find a weight matrix \( W \) that minimizes the difference between these two spaces. This can be formulated as:

\begin{equation}
E(T, M) = \min_W \left( \| T - W^T M \|_F^2 + \delta \| WT - M \|_F^2 \right)
\end{equation}

Here, \( \| \cdot \|_F \) denotes the Frobenius norm, and \( \delta \) is a regularization parameter.

To solve for \( W \), we differentiate \( E(T, M) \) with respect to \( W \):

\begin{equation}
\frac{d E(T, M)}{d W} = -2M (T - W^T M)^T + 2\delta (WT - M) T^T
\end{equation}

Setting the derivative to zero for optimization, we get:

\begin{equation}
-MT^T + MM^T W + \delta WT T^T - \delta MT^T = 0
\end{equation}

Rearranging terms, we derive the modified Sylvester equation:

\begin{equation}
(MM^T) W + W (\delta TT^T) = (1 + \delta) MT^T
\end{equation}

This equation takes the form \( PX + XQ = Z \), where \( P = MM^T \), \( Q = \delta TT^T \), and \( Z = (1 + \delta) MT^T \). The solution \( W \) to this equation is the optimal weight matrix that aligns the Twitter and MHDA-Kb embedding spaces.

By solving the Sylvester equation, we obtain a discriminative weight matrix that effectively modulates word embeddings. This allows the model to dynamically handle domain-specific language and adapt to new terms. This enhanced embedding space significantly improves classification accuracy in mental health contexts, providing a robust framework for analyzing social media content.

\section{Ablation Study}\label{sec:appen_ablation}

\begin{table*}[!htbp]
    \centering
    \begin{tabular}{|p{7cm}|c|c|c|c|c|}
        \hline
        \textbf{Example} & \textbf{A0 (90\%)} & \textbf{A1 (60\%)} & \textbf{A2 (35\%)} & \textbf{A3 (20\%)} & \textbf{A4 (8\%)} \\
        \hline
        \multicolumn{6}{|c|}{Error rates for different analyses, where examples marked with a checkmark (\checkmark) indicate correct classification.} \\
        \hline
        \scriptsize{Just received my COVID-19 test results, and I'm relieved to say they came back negative. \#StaySafe} & & \checkmark & \checkmark & \checkmark & \checkmark \\
        \hline
        \scriptsize{Feeling totally burned out from remote working. \#WorkFromHome \#Stress} & & & & \checkmark & \checkmark \\
        \hline
        \scriptsize{Really feeling the blues with all this isolation. \#PandemicLife \#Mood} & & & & & \checkmark \\
        \hline
        \scriptsize{Worried about my kids acting out so much lately. \#Parenting} & & & & & \checkmark \\
        \hline
    \end{tabular}
    \caption{Classification accuracy improvements from no KBs (A0) to domain-specific LMs (A4) demonstrate the effectiveness of each analysis method.}
    \label{tab:example_classification}
\end{table*}

The ablation study assesses the impact of each component in our analytical framework on reducing error rates, utilizing both quantitative and qualitative analyses. This structured evaluation helps us identify crucial elements for enhancing the performance and reliability of our system.

\subsection{Knowledge-Based}
Our framework leverages diverse knowledge sources to enrich mental health and addiction-related discussion analysis. Employing both general-purpose and domain-specific KBs ensures a comprehensive assessment.

\textbf{Without Knowledge Bases (A0):}
Operating without knowledge bases, our system relies solely on raw text analysis, resulting in higher error rates due to a lack of contextual understanding. This baseline underscores the fundamental necessity of incorporating KBs to reduce errors effectively.

\textbf{With General Purpose KBs (A1):}
General-purpose Kbs help bridge gaps when domain-specific information is scarce. While effective at contextualizing general terms, they often miss the language of mental health discussions, as illustrated below:

\begin{itemize}
    \item \textbf{Green Examples:}
        \begin{itemize}
            \item ``Just received my COVID-19 test results, and I'm relieved to say they came back negative. \#StaySafe"
            \item ``Working from home has become the new normal during the pandemic. Grateful for the flexibility it offers. \#RemoteWork"
        \end{itemize}
    \item \textbf{Red Examples:}
        \begin{itemize}
            \item ``Feeling totally burned out from remote working. \#WorkFromHome \#Stress"
            \item ``Can't sleep at night thinking about how things are going. \#Insomnia \#Anxiety"
        \end{itemize}
\end{itemize}

\textbf{With Domain-Specific KBs (A2):}
Domain-specific KBs refine our model's focus on relevant discussions by employing sophisticated matching techniques beyond simple string matches. This reduces error rates further but still faces challenges with highly contextual language:

\begin{itemize}
    \item \textbf{Green Examples:}
        \begin{itemize}
            \item ``Seeking professional help for my anxiety and depression during the COVID-19 pandemic. Taking steps towards self-care and recovery. \#MentalHealthAwareness \#socialisolation"
            \item ``Supporting loved ones in addiction recovery during these challenging times. Let's break the stigma and offer compassion. \#Addiction \#COVID19"
        \end{itemize}
    \item \textbf{Red Examples:}
        \begin{itemize}
            \item ``Really feeling the blues with all this isolation. \#PandemicLife \#Mood"
            \item ``Worried about my kids acting out so much lately. \#Parenting \#Stress"
        \end{itemize}
\end{itemize}

\subsection{Language Models}
Transitioning to language models addresses the limitations of KBs that rely on direct string matches.

\textbf{With Pre-Trained LMs (A3):}
Pre-trained language models (word2vec) provide broad language comprehension but often miss context-specific nuances of mental health discourse:

\begin{itemize}
    \item \textbf{Green Examples:}
        \begin{itemize}
            \item ``Feeling overwhelmed by the lockdown but staying positive. \#MentalHealth"
        \end{itemize}
    \item \textbf{Red Examples:}
        \begin{itemize}
            \item ``Down in the dumps today because of all the bad news. \#FeelingBlue"
        \end{itemize}
\end{itemize}

\textbf{With Improved and Fine-Tuned LMs (A4):}
Fine-tuning language models on domain-specific data substantially enhances accuracy, reducing misclassifications and improving contextual understanding.

The crucial role of progressively sophisticated KBs and fine-tuning language models in reducing error rates and enhancing the ability to interpret complex social media content related to mental health and drug abuse is illustrated in Tab.~\ref{tab:example_classification}.


\end{document}